%% file: main.tex
\pdfoutput=1

\documentclass[11pt]{article}

\usepackage[preprint]{acl}

\usepackage{times}
\usepackage{latexsym}

\usepackage[T1]{fontenc}

\usepackage[utf8]{inputenc}

\usepackage{microtype}

\usepackage{inconsolata}

\usepackage{graphicx}

\usepackage{booktabs}
\usepackage{multirow}
\usepackage{colortbl}
\usepackage{arydshln}

\usepackage{amsmath}
\usepackage{amssymb}
\usepackage{bbm}
\usepackage{bm}
\DeclareMathOperator*{\argmax}{arg\,max}

\newcommand{\Y}{\mathcal{Y}}
\newcommand{\X}{\mathcal{X}}

\newcommand{\D}{\mathcal{D}}

\usepackage{adjustbox}
\usepackage{subcaption}

\usepackage{xcolor}
\definecolor{backgray}{rgb}{0.9,0.9,0.9}
\definecolor{textgray}{rgb}{0.6,0.6,0.6}
\definecolor{diffgreen}{HTML}{00b06b}
\definecolor{diffred}{HTML}{ff4b00}

\usepackage{tcolorbox}
\usepackage{listings}
\newcolumntype{P}[1]{>{\raggedright\arraybackslash}p{#1}}
\newcommand{\PromptPair}[2]{%
  #1\par
  \vspace{2pt}%
  \begingroup
    \setlength{\fboxsep}{6pt}%
    \colorbox{gray!10}{%
      \parbox{\dimexpr\linewidth-2\fboxsep\relax}{\ttfamily\small #2}%
    }%
  \endgroup
}

\title{Distilling Many-Shot In-Context Learning into a Cheat Sheet}

\author{
  Ukyo Honda\qquad Soichiro Murakami\qquad Peinan Zhang\\
  CyberAgent, Tokyo, Japan\\
  \texttt{\{honda\_ukyo,murakami\_soichiro,zhang\_peinan\}@cyberagent.co.jp}
}

\begin{document}
\maketitle
\begin{abstract}
Recent advances in large language models (LLMs) enable effective in-context learning (ICL) with many-shot examples, but at the cost of high computational demand due to longer input tokens.
To address this, we propose cheat-sheet ICL, which distills the information from many-shot ICL into a concise textual summary (\emph{cheat sheet}) used as the context at inference time.
Experiments on challenging reasoning tasks show that cheat-sheet ICL achieves comparable or better performance than many-shot ICL with far fewer tokens, and matches retrieval-based ICL without requiring test-time retrieval.
These findings demonstrate that cheat-sheet ICL is a practical alternative for leveraging LLMs in downstream tasks.\footnote{The code is publicly available at \url{https://github.com/CyberAgentAILab/cheat-sheet-icl}.}
\end{abstract}

\section{Introduction}
\label{sec:intro}

In-context learning \citep[ICL;][]{brown2020language} has emerged as a novel paradigm for leveraging large language models (LLMs) on downstream tasks \citep{dong-etal-2024-survey}.
Unlike the predominant fine-tuning approach, ICL does not involve updating the model parameters.
Instead, it provides a few task-specific examples, known as \textbf{demonstrations}, and a test input together, enabling LLMs to infer based on this context.
Due to context window limitations, ICL has been typically used in few-shot settings.

Building on the extended context windows afforded by recent advancements in LLMs, \citet{agarwal2024manyshot} and \citet{bertsch-etal-2025-context} have demonstrated the superior performance of \textbf{many-shot ICL}, wherein a larger number of demonstrations are provided, over the conventional few-shot setup.
Although many-shot ICL requires a larger number of demonstrations, it retains key advantages over fine-tuning: it is training-free and can be applied to state-of-the-art proprietary models, which often limit or do not support fine-tuning.
Nevertheless, this paradigm introduces the increased computational costs associated with providing substantially longer input contexts.

\begin{figure}[t]
    \centering
    \includegraphics[width=1.0\linewidth]{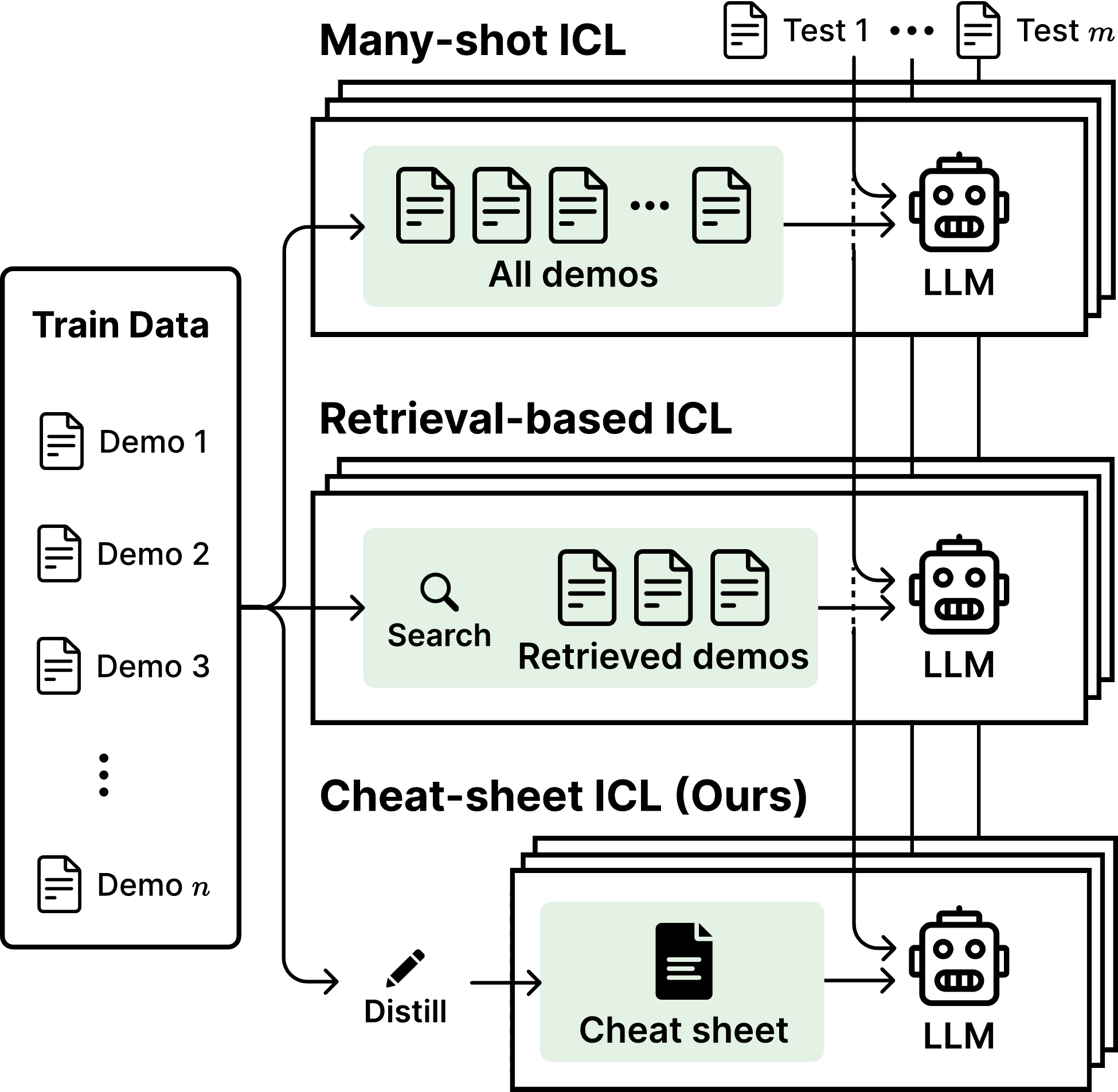}
    \caption{
    Overview of many-shot, retrieval-based, and our cheat-sheet ICL.
    Many-shot ICL processes numerous demonstrations at every inference; retrieval-based ICL involves storing and searching them.
    Cheat-sheet ICL uses a compact cheat sheet distilled from demonstrations, which only needs to be created once.
    }
    \label{fig:overview}
\end{figure}

\looseness=-1
A common approach to efficient ICL, given a large pool of demonstrations, is demonstration retrieval, in which demonstrations are selected based on their similarity to test inputs \citep{liu-etal-2022-makes,bertsch-etal-2025-context}.
While effective, this method requires a retrieval operation for each inference.

In this study, we explore an alternative to existing approaches.
Our key idea is that LLMs, with their advanced language understanding, can extract the knowledge needed for a task as a textual summary.
Rather than requiring LLMs to infer hidden patterns from demonstrations at each inference, we propose distilling many-shot ICL knowledge into an explicit textual format.
Figure~\ref{fig:overview} presents an overview of our method.
We term our approach \textbf{cheat-sheet ICL}, by analogy to how students summarize key points on a single sheet for exams.

Experimental results show that this remarkably simple method achieves performance comparable to, or even surpassing, many-shot ICL on challenging reasoning tasks.
Furthermore, cheat sheets are interpretable and allow easy intervention for further improvements.
Moreover, cheat-sheet ICL matches demonstration retrieval in performance, indicating its viability as an alternative.
We consider this alternative paradigm a promising direction, and anticipate further improvements as LLMs' language understanding advances.

\section{Preliminaries}
\label{sec:preliminaries}

\subsection{ICL and Many-Shot ICL}
\label{sec:many-shot}
ICL performs inference by conditioning on a set of demonstrations alongside the test input \citep{brown2020language}.
Let $\mathcal{X}$ and $\mathcal{Y}$ denote the input and label spaces, respectively.
We define a set of $n$ demonstrations as $\D_{n} := \{(x_i, y_i)\}_{i=1}^{n}$, where $x_i \in \X$ represents an input instance and $y_i \in Y$ denotes the corresponding label.
ICL selects the sequence of tokens that maximizes the conditional probability given the concatenated demonstrations and test input $x^{\text{test}}$:
\begin{align}
    \label{eq:icl}
    y^{*} = \argmax_{y \in \Y} P(y | \D_n, x^{\text{test}}).
\end{align}

Recent advancements have enabled LLMs to process substantially longer sequences of input tokens, thereby increasing $n$ by orders of magnitude.
This regime, characterized by a large number of in-context demonstrations, is referred to as many-shot ICL \citep{agarwal2024manyshot,bertsch-etal-2025-context}.

\subsection{Improved ICL Baseline}
\label{sec:rationale_aug}
For all variants of ICL, we adopt \emph{reinforced ICL}, which was shown to outperform vanilla ICL across a broad range of shots \citep{agarwal2024manyshot}.
This method augments demonstrations with model-generated rationales, which are obtained by sampling multiple chain-of-thought (CoT) reasoning paths \citep{wei2022chain} from LLMs and selecting the correct paths.

We follow this approach, but we augment the rationales more efficiently, as proposed in X-ICL \citep{he-etal-2024-using}.
X-ICL augments rationales for demonstrations by sampling explanations $\hat{r}$ from LLMs conditioned on both the input and its correct label.
In this way, it is possible to collect rationales that successfully lead to the correct answer with a single sampling.
\emph{Unless otherwise specified, throughout all experiments in this study, all methods use rationale-augmented demonstrations, $\hat{\D}_{n} := \{(x_i, \hat{r}_i, y_i)\}_{i=1}^{n}$}.

\section{Method: Cheat-Sheet ICL}
\label{sec:method}

\looseness=-1
While many-shot ICL performs well across multiple reasoning tasks, its computational cost is much higher than that of conventional few-shot ICL, due to the increased number of input tokens.\footnote{
Although caching previously processed prefixes reduces the prefill cost for repeated many-shot inputs, the computational cost of decoding remains substantial, as attention must still be performed over the long context \citep{bertsch-etal-2025-context}.
Moreover, in hosted API settings, caches are often evicted after short intervals or require paid persistence to retain them.
}
To reduce the inference cost, we propose cheat-sheet ICL, where patterns learned through many-shot demonstrations are summarized in a compact cheat sheet.

The key intuition behind our approach is that LLMs may be able to represent the knowledge they have learned in the form of text, just as humans do.
Recent LLMs have a high level of language understanding and leverage the patterns they have learned in textual CoT reasoning.
Thus, the entire set of learned patterns can potentially be summarized in textual form, eliminating the need to store many-shot demonstrations in the context and extract patterns from them at each inference.

\begin{figure*}[t]
\centering
\includegraphics[width=1.0\textwidth,keepaspectratio]{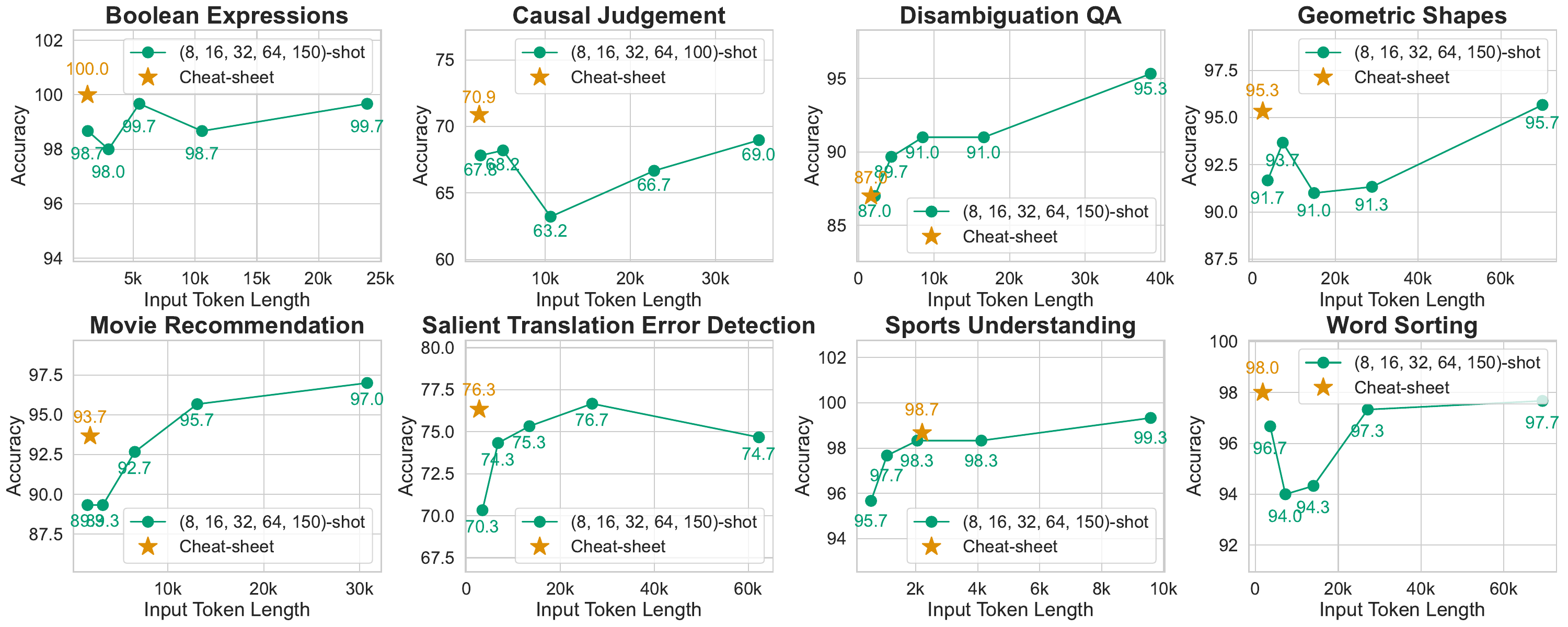}
\caption{
Main results obtained with GPT-4.1. Scores are averaged over three runs.
}
\label{fig:main}
\end{figure*}

\subsection{Cheat-Sheet Creation}
\label{sec:cs_creation}
The essential step in cheat-sheet ICL is the preprocessing step of creating the cheat sheet.
\emph{Note that this preprocessing is executed only once for each task and requires no additional operations during inference.}
Specifically, we present LLMs with the entire set of demonstrations $\hat{\D}_n$, together with a specifically designed prompt, as shown in Appendix~\ref{app:cheat_prompt}.
This prompt is intended to guide LLMs in concisely extracting the core knowledge essential for solving the target task.
We refer to the output as a cheat sheet $S$, drawing an analogy to the way students condense crucial information onto a single sheet of paper to assist in answering exam questions.

\subsection{Inference}
\label{sec:inference}
During inference, we present LLMs with the cheat sheet $S$ and the test input $x^{\text{test}}$.
We do not provide the entire set of $\hat{\D}_n$, but only two examples $\hat{\D}_2 = \{(x_i, \hat{r}_i, y_i)\}_{i=1}^{2}$, as format instructions to guide the LLMs in producing outputs in the desired format.
Formally, the decision making in cheat-sheet ICL is defined as follows:
\begin{align}
    \label{eq:cs-icl}
    y^{*} = \argmax_{y \in \Y} P(y | S, \hat{\D}_2, x^{\text{test}}).
\end{align}

\section{Experiments}
\label{sec:experiments}

We test whether the cheat sheet distilled from many-shot demonstrations can achieve comparable performance in ICL, despite using far fewer tokens.

\subsection{Experimental Setup}
\label{sec:setup}

Our goal is to offer an efficient alternative to many-shot ICL.
To test whether our approach can retain many-shot ICL performance with fewer tokens, we should use datasets in which many-shot ICL outperforms few-shot ICL.
Otherwise, having only a few demonstrations may already be sufficient to perform well on the tasks, or simply reducing the number of tokens may benefit LLMs.
As a preliminary experiment, we ran few-shot and many-shot ICL on all the reasoning tasks tested in \citet{agarwal2024manyshot}: BIG-Bench Hard \citep[\textbf{BBH};][]{suzgun-etal-2023-challenging,srivastava2023beyond}, MATH500 \citep{lightman2024lets}, GSM8K \citep{cobbe2021training}, and GPQA \citep{rein2024gpqa}.
We then selected \emph{eight challenging BBH reasoning tasks, which were the only ones where many-shot outperformed few-shot by more than one percentage point.}
To effectively process up to 250k tokens in our experiments, we were limited to using advanced proprietary models.
Unless otherwise specified, all rationale augmentation, cheat-sheet creation, and inference were performed using \textbf{GPT-4.1}.
See Appendices~\ref{app:setup_details}--\ref{app:model_selection} for further details on the experimental setup and the selection of datasets and models.

\subsection{Main Results}
\label{sec:main_results}
We show the main results in Figure~\ref{fig:main}.
In seven out of the eight tasks, cheat-sheet ICL outperforms few-shot ICL with the same or a smaller input token budget.
Even when compared to many-shot ICL on the far right of the figures, cheat-sheet ICL achieves comparable or even better performance while using far fewer input tokens.\footnote{
The essence of many-shot learning lies not in the number of examples, but in the number of tokens in context.
For example, 1,000 examples in the MNLI dataset amount to only around 45,000 tokens, while the many-shot setting in BBH benchmarks uses about 70,000 tokens with 150 examples.
The scale of many-shot learning thus closely depends on the number of tokens within each example.
}
These results clearly demonstrate the effectiveness and efficiency of cheat-sheet ICL.
We present the time and monetary costs in Appendix~\ref{app:costs}.

Appendices \ref{app:expt_no_aug} and \ref{app:expt_sc} further demonstrate that cheat-sheet ICL remains effective both without the rationale augmentation introduced in Section~\ref{sec:rationale_aug} and when employing the self-consistency decoding algorithm \citep{wang2023selfconsistency}.
Appendix~\ref{app:expt_prompt} shows that modest variations of the cheat-sheet prompt yield comparable downstream performance.
These results underscore the robustness of our approach.

\begin{figure*}[t]
\centering
\includegraphics[width=1.0\textwidth,keepaspectratio]{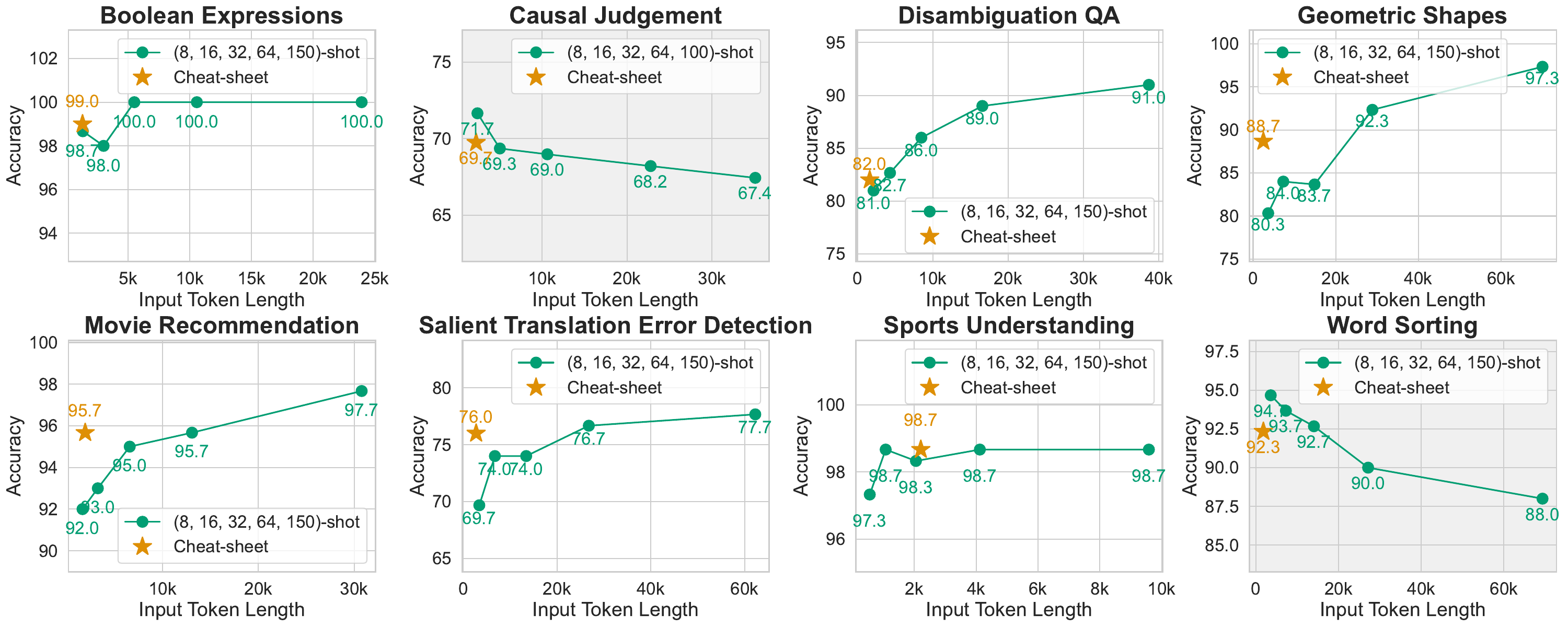}
\caption{
Results on cheat sheet transferability with Gemini 2.0 Flash.
Scores are averaged over three runs.
}
\label{fig:transfer}
\end{figure*}

\subsection{Error Analysis and Interpretability}
\label{sec:error}
Although cheat-sheet ICL performs well overall, it struggles with the Disambiguation QA task, which requires identifying the antecedent of a pronoun or answering ``ambiguous'' if it cannot be logically determined.
We found that cheat-sheet ICL often incorrectly relied on common sense when the answer should be ``ambiguous''.\footnote{For example, interpreting ``their office'' as a director's office, since meetings with a director are more likely to be held there than in the other party's office.}
Because the cheat sheet is human-interpretable, we could easily identify and remove the section that encouraged using world knowledge, and add an explicit instruction not to use it.\footnote{
By contrast, a conventional list of in-context demonstrations offers little insight into which examples influence the model's output or how.
For example, in Figure~\ref{fig:main}, certain mid-shot ICL settings suffer sudden drops in performance on some tasks, but the demonstration list by itself does not make the underlying cause apparent.
}
See Appendix~\ref{app:manual_revision} for the exact modifications we applied.
This simple modification of the cheat sheet improved accuracy from 87.0 to 89.7.
This interpretability and targeted modification are practical advantages of cheat-sheet ICL.

\subsection{Transferability of Cheat Sheets}
\label{sec:transfer}
In addition, we tested whether the generated cheat sheets are also effective when used with another model.
Specifically, we provided \textbf{Gemini 2.0 Flash} with the same cheat sheets created using GPT-4.1.
Figure~\ref{fig:transfer} shows the results.
The cheat sheets yield similar improvements in most cases, confirming their good transferability across models.
The exceptions are those graphs highlighted with the \colorbox{backgray}{gray background}.
Although cheat-sheet ICL underperforms few-shot ICL in these cases, the tasks show no gains from many-shot ICL with this model and thus are not a suitable benchmark for evaluating our method.
At the very least, cheat-sheet ICL still outperforms many-shot ICL in these cases.

\subsection{Comparison with Retrieval Methods}
\label{sec:retrieval}

\looseness=-1
When a large pool of demonstrations is available, retrieving those that are similar to the test inputs is known to be effective \citep{liu-etal-2022-makes}.
We compared cheat-sheet ICL with three retrieval methods: \textbf{BM25}, which uses exact-match search to retrieve demonstrations; \textbf{Cosine}, which retrieves demonstrations based on cosine similarity in the embedding space; and \textbf{Set-BSR}, which uses BERTScore \citep{Zhang2020BERTScore:} to capture various aspects of similarity to test inputs \citep{gupta-etal-2023-coverage}.
We retrieved eight demonstrations, following \citet{gupta-etal-2023-coverage}.
More experimental details are in Appendix~\ref{app:retrieval}.

Table~\ref{tab:retrieve} shows that retrieval-based ICL and cheat-sheet ICL achieve comparable performance on the BBH tasks with similar input token lengths.
A practical benefit of cheat-sheet ICL is that the cheat sheet needs to be created only once per task, and it does not require storing or searching the full demonstration pool during every inference.
Thus, the competitive performance demonstrates that cheat-sheet ICL is an efficient alternative.

\input{tables/retrieve}

\section{Related Work}
\label{sec:realted_work}

\subsection{Efficiency in Many-Demonstration Scenarios}
\label{sec:related_efficiency}
\looseness=-1
Demonstration retrieval has been used to exploit large sets of examples in ICL \citep{liu-etal-2022-makes,luo2024incontext}, and it also benefits many-shot ICL scenarios \citep{bertsch-etal-2025-context}.
Recently, attention modification techniques have been proposed to better highlight important information in many-shot demonstrations \citep{yuan-etal-2024-focused}.
However, demonstration retrieval requires retrieval computations at inference time, and attention modification necessitates access to model parameters, a requirement that is impractical for state-of-the-art proprietary LLMs.
In contrast, cheat-sheet ICL can be applied without additional inference-time computational cost or parameter access.
\citet{wan2025from} found that a small number of influential demonstrations can match the performance of full many-shot settings.
However, they did not explore how to leverage this insight for efficiency, focusing instead on increasing the number of influential examples to the many-shot scale.

\subsection{Instruction Induction and Prompt Compression}
\label{sec:related_inst}
Automatic instruction induction from few-shot demonstrations--including iterative variants on small subsets--predates the many-shot regime, but these methods were not designed for efficiency under many-shot ICL \citep{honovich-etal-2023-instruction,zhou2023large}.
Meanwhile, prompt compression has largely focused on shrinking RAG inputs or lengthy knowledge sources rather than demonstration sets, and often relies on costly architectural/parameter changes or on iterative optimization over small subsets \citep{li-etal-2025-prompt}.
In contrast, we study the many-shot capabilities of recent LLMs and investigate whether the knowledge learned from numerous demonstrations can be distilled, in a single pass, into a compact cheat sheet, without additional training or model modifications.
Accordingly, we evaluate our method against many-shot ICL and empirically demonstrate its effectiveness, unlike prior work focused on zero-/few-shot benchmarks.

\subsection{Knowledge Distillation}
\label{sec:distill}
Knowledge distillation aims to transfer knowledge from a large teacher model to a smaller student model.
\citet{hinton2015distilling} proposed training the student model to match the teacher's output probabilities, while \citet{west-etal-2022-symbolic} relaxed the need for probability outputs by instead using the teacher's output texts.
In contrast, cheat-sheet ICL encodes task-specific knowledge into a concise textual summary rather than model parameters, making it applicable to adapting proprietary LLMs to specific tasks.

\section{Conclusion}
\label{sec:conclusion}

We introduced cheat-sheet ICL, which uses concise textual summaries distilled from many-shot demonstrations to leverage LLMs.
This approach matched or exceeded the performance of many-shot ICL and retrieval methods on challenging reasoning tasks, while remaining more efficient and interpretable.
We believe that cheat-sheet ICL provides a simple and practical alternative for leveraging LLMs.

\section*{Limitations}
\label{sec:limitations}

In this study, we limit our focus to reasoning tasks, as they involve explicit use of knowledge in text and thus provide a good testbed for our method.
Future work will extend our method to tasks where explicit and comprehensive reasoning is less central, \emph{e.g.}, dialogue generation and creative writing.
That said, even in creative settings such as advertising, text generation exhibits recurring conventions aligned with human preferences, such as using bracketed tokens for emphasis and favoring noun-dense phrasing \citep{murakami-etal-2025-adparaphrase,murakami-etal-2025-adparaphrase-v2}.
We therefore expect our cheat-sheet ICL approach to remain effective by identifying and leveraging such patterns in the provided examples.

As discussed in Section~\ref{sec:transfer}, our method does not show clear improvements on tasks where many-shot demonstrations do not outperform few-shot settings.
However, this is a limitation of the many-shot ICL paradigm itself, rather than our method.
Since some datasets are solved well by LLMs with only a few demonstrations, providing many examples may even distract the model from the required output format (see Appendix~\ref{app:academic}).
So far, there is no precise way to identify tasks for which many-shot ICL is more suitable than few-shot ICL.
Therefore, we recommend conducting a preliminary check on a small subset of data to see whether many-shot ICL or cheat-sheet ICL performs better than few-shot ICL when addressing a new task.

Our approach requires computing many-shot ICL outputs as preprocessing for cheat-sheet creation.
However, note that this needs to be done only once per task, unlike many-shot ICL, which incurs this cost for every test input.

\looseness=-1
Many-shot and cheat-sheet ICL methods leverage long-context capabilities, and our experiments required processing inputs up to about 250,000 tokens.
Currently, only advanced proprietary models can effectively handle such long contexts, so our evaluation is limited to GPT-4.1 and Gemini 2.0 Flash.
The effectiveness of our method with these state-of-the-art models suggests wider applicability, as we expect that future open-source models will also be able to effectively comprehend and utilize similarly long contexts.
While the nondeterminism of these proprietary models introduces modest variance in performance, it is noteworthy that the cheat-sheet construction stage does not increase variance beyond that observed in few-shot or many-shot ICL in most cases; see Appendix~\ref{app:proprietary_var} for details.

Interpretability is limited to what is explicitly stated in the cheat sheets.
When a cheat sheet is oversimplified and the LLM reverts to prior knowledge not covered there, the LLM's failure cases can be hard to understand by examining the cheat sheet alone.
Encouraging more detailed cheat sheets through prompt engineering could be a possible direction to enhance interpretability.

It remains unclear under what conditions cheat-sheet ICL can be as effective as many-shot ICL.
Based on its consistent failure to match many-shot ICL on Disambiguation QA, together with our error analysis, we speculate that rules intended to override commonsense priors are particularly difficult for LLMs to induce as admissible constraints.
Commonsense reasoning is likely reinforced during pretraining, making its suppression unnatural for the model.
That said, requirements to ignore common sense are uncommon outside benchmark datasets, so we expect the practical risk of cheat-sheet ICL underperforming in real-world applications to be limited.

While our study empirically demonstrates the effectiveness of cheat-sheet ICL, we do not pursue a formal theoretical treatment.
We anticipate that our results will spur further theoretical work, for example, quantifying how aggressively many-shot demonstrations can be compressed while maintaining a specified level of performance.

\section*{Ethical Considerations}
\label{sec:ethical}

We do not foresee any ethical issues arising specifically from our method.
All the datasets we used are publicly available and commonly used for research purposes.

\bibliography{custom}

\clearpage
\appendix

\section{Prompt for Cheat-Sheet Creation}
\label{app:cheat_prompt}

We employed the following prompt to create a cheat sheet for each task.
Note that $\hat{\D}_n$ varies by task.
The prompt was fed to GPT-4.1, and the output was used as the cheat sheet $S$.

\begin{tcolorbox}[title=Prompt for Cheat-Sheet Creation, colback=gray!10, colframe=black!70, boxrule=0.5pt, top=0pt, bottom=0pt, left=0pt, right=0pt, arc=0.8mm]
Create a cheat sheet based on the examples below. You will be asked to answer questions similar to these examples during the test, without being allowed to refer to the examples at that time. Your task here is to make a cheat sheet that will help you answer such problems correctly. First, carefully read the examples below and identify which ones you find most difficult to answer.\\\\
\{$\hat{\D}_n$\}\\\\
Now, create a cheat sheet to help you solve the difficult examples. Exclude any content that is easy for you, and only include specific, detailed points to address the challenging ones.
\end{tcolorbox}

\section{Further Setup Details}
\label{app:setup_details}

\subsection{Datasets}
\label{app:datasets}
As described in Section~\ref{sec:setup}, we selected datasets in which many-shot ICL outperforms few-shot ICL.
This selection allows us to accurately evaluate whether our method preserves many-shot performance with far fewer tokens, by avoiding datasets for which only a few demonstrations may be sufficient to perform well on the tasks, or where simply reducing the number of tokens may benefit LLMs.

For the BBH benchmark,\footnote{\url{https://github.com/suzgunmirac/BIG-Bench-Hard}} we conducted single runs and selected tasks for which the full many-shot setting (using either 100 or 150 shots, depending on the specific task) outperformed the 8-shot setting by more than one percentage point in accuracy.
This criterion resulted in a selection of the eight tasks: Boolean Expressions, Causal Judgement, Disambiguation QA, Geometric Shapes, Movie Recommendation, Salient Translation Error Detection, Sports Understanding, and Word Sorting.

Except for Causal Judgement, each dataset consists of 250 instances.
We adopted the train--test splits from \citet{agarwal2024manyshot}, allocating 150 examples as demonstrations for ICL and using the remaining 100 examples as the test set.
For Causal Judgement, due to its smaller dataset size, we used 100 examples for demonstrations and the remaining 87 examples for evaluation.

We also applied the same data selection criterion to academic reasoning tasks, namely MATH500, GSM8K, and GPQA, in order to comprehensively cover the range of reasoning benchmarks examined in \citet{agarwal2024manyshot}.
However, we found that none of these datasets satisfied the selection threshold.
Further details are provided in Appendix~\ref{app:academic}.

All the datasets we used are in English and publicly available for research purposes.

\subsection{Task Descriptions}
\label{app:tasks}
Below are brief descriptions of the eight selected reasoning tasks \citep{suzgun-etal-2023-challenging,srivastava2023beyond}.

\begin{itemize}
\item \textbf{Boolean Expressions:} Answer True or False given a sequence of boolean constants (True or False) and boolean operations (and, or, not).
\item \textbf{Causal Judgment:} Answer Yes or No if a typical person would answer the question about causation in the way provided.
\item \textbf{Disambiguation QA:} Answer which antecedent a pronoun refers to, or answer ``ambiguous'' if it cannot be logically determined.
\item \textbf{Geometric Shapes:} Identify the geometric shape of an SVG path element.
\item \textbf{Movie Recommendation:} Select which movie in a list is similar to another list of movies.
\item \textbf{Salient Translation Error Detection:} Indicate which type of translation error can be detected in a German--English translation.
\item \textbf{Sports Understanding:} Answer yes or no if a given sentence relating to sports is plausible.
\item \textbf{Word Sorting:} Sort a list of words in alphabetical order.
\end{itemize}

\subsection{Rationale Augmentation}
\label{app:rationale_aug_details}
As described in Section~\ref{sec:rationale_aug}, all ICL methods used rationale-augmented demonstrations unless otherwise specified.
For rationale augmentation in the BBH tasks, we used the CoT prompts prepared for each task, each consisting of three demonstrations with human-annotated rationales: $\{(x'_j, r'_j, y'_j)\}_{j=1}^{3}$.
To align with X-ICL's rationale generation format, which is called \textbf{meta-prompt} in \citet{he-etal-2024-using}, we format them as follows:

\begin{tcolorbox}[title=Format of Meta-Prompt, colback=gray!10, colframe=black!70, boxrule=0.5pt, top=0pt, bottom=0pt, left=0pt, right=0pt, arc=0.8mm]
Question: \{$x'_1$\}\\
Answer: \{$y'_1$\}\\
Explanation: \{$r'_1$\}\\
\#\#\#\\
Question: \{$x'_2$\}\\
Answer: \{$y'_2$\}\\
Explanation: \{$r'_2$\}\\
\#\#\#\\
Question: \{$x'_3$\}\\
Answer: \{$y'_3$\}\\
Explanation: \{$r'_3$\}\\\#\#\#\\
Question: \{$x_i$\}\\
Answer: \{$y_i$\}\\
Explanation:
\end{tcolorbox}

We provided the LLM with the meta-prompt combined with each input--answer pair of demonstrations $(x_i, y_i) \in \D_n$, and used the output as the augmented rationale $\hat{r}$ for each demonstration.

For the academic tasks, we constructed the meta-prompt by selecting the first three training examples that included human-annotated rationales.
The subsequent procedure remained identical to that used for the BBH tasks.

\subsection{Models}
\label{app:models}
The specific version of GPT-4.1 used was \texttt{gpt-4.1-2025-04-14}, and the version of Gemini 2.0 Flash was \texttt{gemini-2.0-flash-001}.
The models were accessed via the Azure OpenAI API and the Gemini API, respectively.

\subsection{Decoding Configurations}
\label{app:decoding}
We set the temperature to 0 to maximize reproducibility.
Deterministic decoding is not available for the proprietary models we used.
To ensure that outputs conformed to the format shown in the demonstrations, we used the following system prompt:

\begin{tcolorbox}[title=System Prompt, colback=gray!10, colframe=black!70, boxrule=0.5pt, top=0pt, bottom=0pt, left=0pt, right=0pt, arc=0.8mm]
Answer the question by following the provided examples. Ensure that your response ends with Answer: and your final answer.
\end{tcolorbox}

\begin{figure*}[t]
\centering
\includegraphics[width=1.0\textwidth,keepaspectratio]{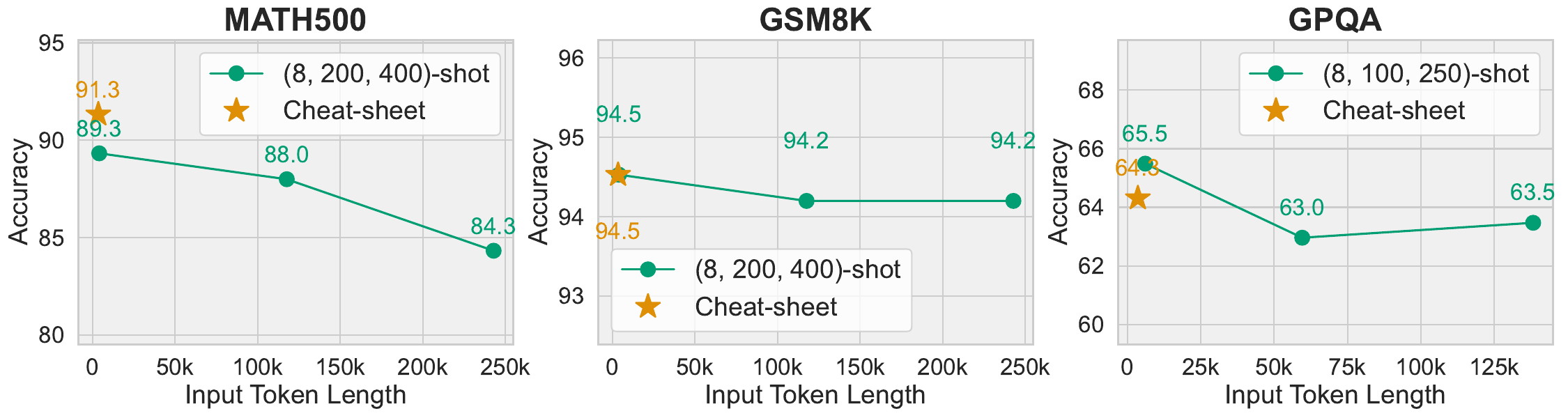}
\caption{
Results on academic tasks obtained with GPT-4.1.
Scores are averaged over three runs.
}
\label{fig:acad}
\end{figure*}

\subsection{Evaluation}
\label{app:eval}
All the tasks are evaluated using accuracy.
Results are averaged over three runs with different random seeds, which affect the ordering of training data.
For the vanilla ICL, demonstrations are selected as the first $n$ examples based on the shuffled data from each seed, resulting in different demonstrations across seeds.
In the full many-shot setting, the demonstration set remains the same, but the order varies by seed.
For cheat-sheet ICL, a cheat sheet is created from the demonstrations reordered according to each random seed.
For the format-instruction examples $\hat{\D}_2$ in Eq.~\eqref{eq:cs-icl}, we select the first two from the reordered demonstrations.
In retrieval-based ICL, the retrieved demonstrations are also reordered based on the seed.

For token counting, we employed the OpenAI \texttt{tiktoken} with the \texttt{o200k\_base} encoding.\footnote{\url{https://github.com/openai/tiktoken}}

\section{Dataset Selection on Academic Tasks}
\label{app:academic}
As described in Section~\ref{sec:setup} and Appendix~\ref{app:datasets}, we also conducted our dataset selection on two mathematical datasets, MATH500 and GSM8K, as well as GPQA, a multiple-choice QA dataset spanning the domains of biology, physics, and chemistry.
However, none of these datasets showed improvements under many-shot ICL and therefore did not meet our selection criterion.

\subsection{Setup}
\label{app:setup_academic}
Our experimental setup adheres to that of \citet{agarwal2024manyshot}.
However, the MATH dataset used for demonstrations in the mathematical tasks is currently unavailable due to copyright restrictions.\footnote{\url{https://huggingface.co/datasets/hendrycks/competition_math/discussions/5}}
To approximate the experimental conditions of \citet{agarwal2024manyshot}, we partitioned the MATH500 dataset,\footnote{\url{https://huggingface.co/datasets/HuggingFaceH4/MATH-500}} using 400 examples as demonstrations and reserving the remaining 100 examples for testing.
For GSM8K,\footnote{\url{https://huggingface.co/datasets/openai/gsm8k}} we evaluated 500 examples from the test split in a transfer setting, wherein the demonstration examples are drawn from MATH500, as described above.
For GPQA,\footnote{\url{https://huggingface.co/datasets/Idavidrein/gpqa}} we used the \texttt{gpqa\_diamond} split as the test set, and constructed the demonstration set from the non-overlapping instances in \texttt{gpqa\_main}.
Consequently, the GPQA test set comprised 198 instances, while the demonstration set contained 250 examples.

All the tasks were evaluated based on accuracy.
For mathematical tasks, which require careful parsing of model outputs, we followed the publicly available evaluation script released by OpenAI.\footnote{\url{https://github.com/openai/simple-evals}}

\subsection{Results}
\label{app:results_academic}
The results are provided in Figure~\ref{fig:acad}.
Contrary to expectations, we observe no performance improvement when increasing the number of in-context demonstrations from few-shot to many-shot.
One plausible interpretation is that the academic knowledge assessed by these benchmarks is sufficiently general and already robustly encoded in the state-of-the-art LLM GPT-4.1, such that a small number of demonstrations is sufficient to elicit the relevant capabilities.
For example, although few-shot math demonstrations cannot encompass all possible problem types, the model often answers correctly, presumably by drawing on prior knowledge acquired during pretraining.
Under these conditions, adding more demonstrations may yield diminishing returns, providing little or no additional benefit.

Notably, the many-shot setting decreased performance relative to the few-shot baseline.
In our analysis of model outputs, we found that the many-shot setting led to a higher frequency of output format errors compared to the few-shot setting.
We attribute the performance degradation in part to such errors, potentially caused by the increased length of the input distracting the model from adhering to the required answer format.

Cheat-sheet ICL did not yield improvements over few-shot ICL, but outperformed many-shot ICL.
These findings are in line with the results highlighted in gray in Figure~\ref{fig:transfer}.
Note that our goal is to offer an efficient alternative to many-shot ICL; we do not aim to resolve the failure modes of many-shot ICL, particularly on tasks where it underperforms few-shot ICL.
Even on these tasks, cheat-sheet ICL substantially reduces input length relative to many-shot ICL while still outperforming it, demonstrating that the intended improvements were achieved.

In contrast to these academic knowledge benchmarks, BBH is more oriented toward pattern recognition within datasets, rather than being a testbed for general knowledge.
The use case for many-shot ICL should often be adaptation to specific tasks, rather than general tasks for which LLMs are already well pretrained.
This makes BBH a more suitable benchmark for evaluating many-shot ICL with recent stronger LLMs.

\input{tables/acad_gemini}

\input{tables/cost}

\section{Model Selection}
\label{app:model_selection}
We used GPT-4.1 in our experiments instead of \texttt{gemini-1.5-pro-001}, which was the only model evaluated as a many-shot learner in the original many-shot ICL paper \citep{agarwal2024manyshot}, for the following reasons.

First, \texttt{gemini-1.5-pro-001} is no longer available, and its results are not reproducible.
The closest available alternative is \texttt{gemini-1.5-pro-002}, so we evaluated whether this model could achieve better many-shot performance than few-shot on academic benchmarks, where GPT-4.1 could not (see Appendix~\ref{app:academic}).
As shown in Table~\ref{tab:gemini_academic}, \texttt{gemini-1.5-pro-002} achieves marginally better many-shot performance than few-shot on GSM8K, but the results on the other benchmarks are nearly identical: the many-shot setting does not yield better performance than the few-shot setting.
We also observe that our cheat-sheet ICL matches or surpasses many-shot ICL, consistent with our GPT-4.1 results.
As discussed in Appendix~\ref{app:academic}, we speculate that the lack of gains from increasing the number of shots is because the academic knowledge assessed by these benchmarks is sufficiently general and has already been robustly acquired as prior knowledge by recent state-of-the-art LLMs.

Second, GPT-4.1 is a sufficiently strong many-shot learner in our experimental settings.
The original many-shot ICL paper reported that \texttt{gemini-1.5-pro-001} achieved better performance in the many-shot setting compared to the few-shot setting across eight BBH datasets.
Similarly, we found that GPT-4.1 also showed improved many-shot performance on the same number of eight BBH datasets.
The only exception is the academic tasks, for which, as shown above, \texttt{gemini-1.5-pro-002} also does not exhibit many-shot gains.

Finally, we selected GPT-4.1 to better simulate practical use cases.
GPT-4.1 is one of the most advanced LLMs and substantially outperforms \texttt{gemini-1.5-pro-001} on standard LLM benchmarks while maintaining a similar cost per token.\footnote{See, for example, \url{https://artificialanalysis.ai/} for comparative benchmark results.}
From a practical perspective, methods should be applicable to models that offer a better performance--cost trade-off.
This provided sufficient justification for us to test our method with GPT-4.1 rather than \texttt{gemini-1.5-pro-001}.

\begin{figure*}[t]
\centering
\includegraphics[width=1.0\textwidth,keepaspectratio]{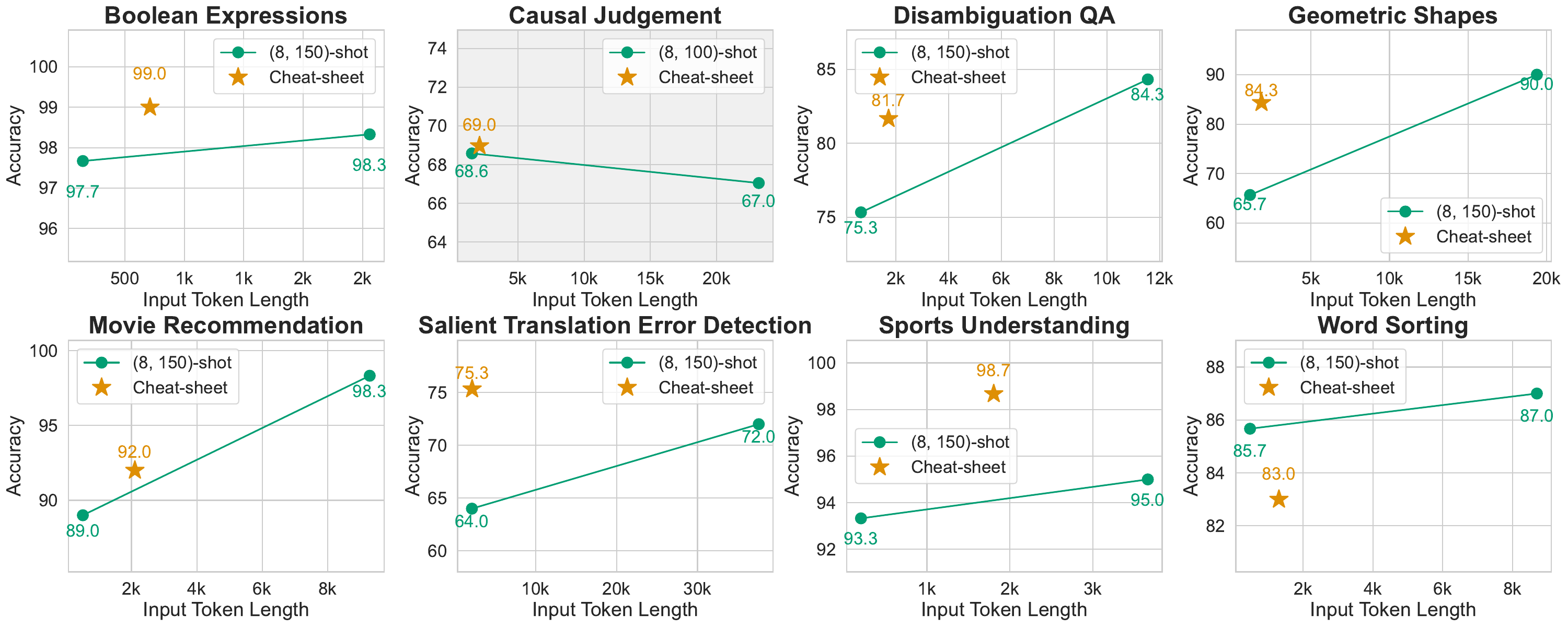}
\caption{
Performance of GPT-4.1 without rationale augmentation in the few-shot, many-shot, and cheat-sheet ICL settings. Scores are averaged over three runs.
}
\label{fig:no_aug}
\end{figure*}

\begin{figure*}[t]
\centering
\includegraphics[width=1.0\textwidth,keepaspectratio]{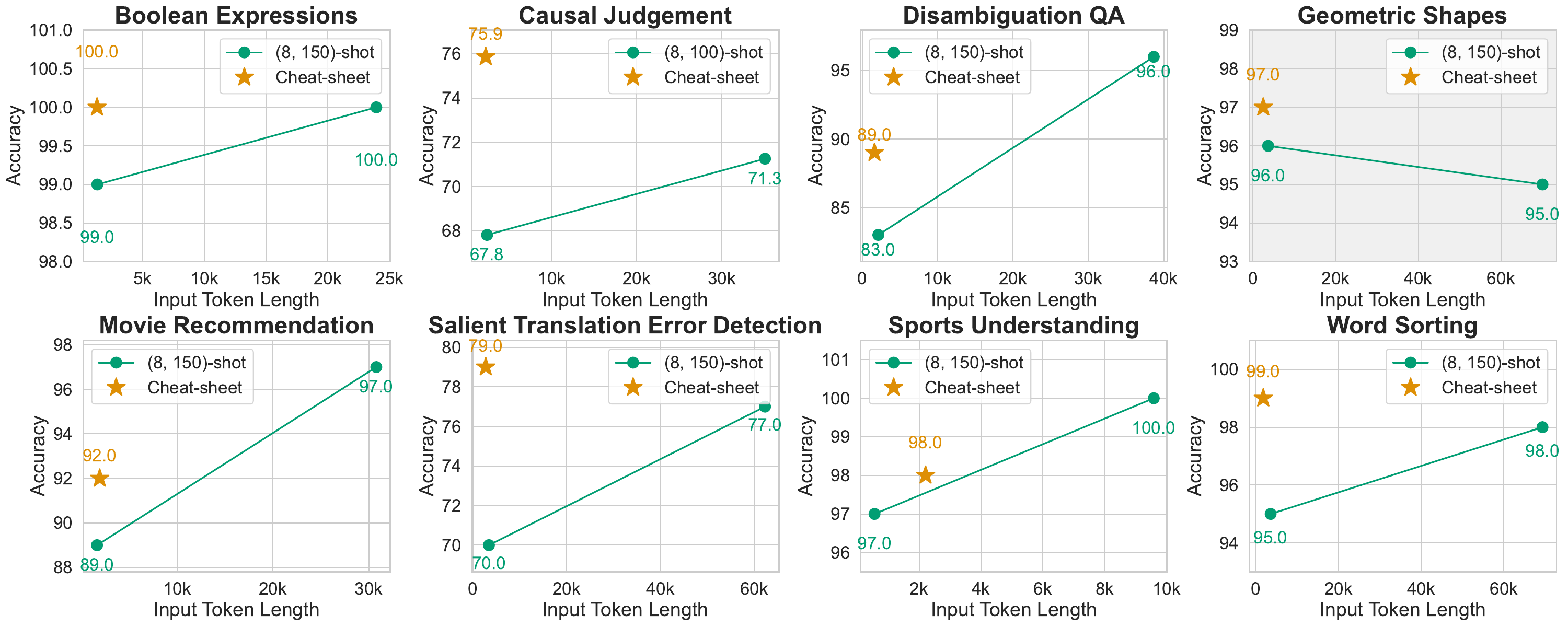}
\caption{
Performance of GPT-4.1 with self-consistency decoding in the few-shot, many-shot, and cheat-sheet ICL settings. Scores are averaged over three runs.
}
\label{fig:sc}
\end{figure*}

\input{tables/prompt_var_desc}
\input{tables/prompt_var_results}

\input{tables/proprietary_var}

\section{Time and Monetary Cost}
\label{app:costs}
We also report the wall-clock time and monetary cost, both of which are closely tied to token length.
All results are from the Boolean Expressions task, evaluated with the same random seed.
All runs used the Azure OpenAI API's prompt caching, which reuses previously processed prefixes.
Accordingly, we computed monetary cost using GPT-4.1 cached-token pricing (as of May 2025).

Table~\ref{tab:costs} shows the results.
As indicated by the average token length, output lengths differ only slightly across settings, whereas input lengths differ substantially: the 150‑shot setting uses much longer inputs.
Accordingly, 150‑shot ICL incurs a substantially higher input‑encoding cost.
Even with caching of the repeated many-shot inputs, 
150-shot ICL remains markedly slower in wall-clock time, plausibly because decoding requires attending to the long cached context at each generation step.
By contrast, cheat-sheet ICL closely matches 8-shot ICL in both time and cost owing to similar token lengths, while retaining the strong performance of 150-shot ICL.

\section{Experiment without Rationale Augmentation}
\label{app:expt_no_aug}
Except for the ablation study in this section, all experiments used rationale augmentation (see Section~\ref{sec:rationale_aug}), which has been reported to improve many-shot ICL \citep{agarwal2024manyshot}.
We generated rationales automatically using GPT-4.1, starting from only a handful of manually annotated seed rationales.
However, sampling rationales for all demonstrations can sometimes be computationally intensive.
We therefore test whether our method remains effective without rationale augmentation.

Figure~\ref{fig:no_aug} shows the results.
Our cheat-sheet ICL largely matches the performance of many-shot ICL while requiring far fewer input tokens, and even outperforms it on half of the datasets, demonstrating the robustness of our approach.

We also observe that performance with rationale augmentation in our main results is noticeably higher than in the setting without augmentation.
Given that the augmentation relied on only three manually annotated seed rationales, these findings further confirm the practical benefit of employing rationale augmentation.

\section{Experiment with Self-Consistency}
\label{app:expt_sc}
Following the experimental settings of \citet{agarwal2024manyshot}, we approximated greedy decoding in our main experiment by setting the temperature to 0 \citep{team2024gemini}.
To further assess the effectiveness of our method under different decoding algorithms, we conducted experiments using self-consistency (SC) decoding \citep{wang2023selfconsistency}, which is known to be a strong but computationally expensive approach.

SC samples multiple responses at a relatively high temperature (typically 0.7) and then selects the most frequent answer via majority voting.
We evaluated few-shot, many-shot, and cheat-sheet ICL under SC.
In our experiments, we set the temperature to 0.7 and sampled three responses per input, resulting in approximately three times higher decoding cost than greedy decoding.

Figure~\ref{fig:sc} shows the results.
The SC performance differs slightly from that in the main experiment.
Nevertheless, the effectiveness of our cheat-sheet ICL remains the same: cheat-sheet ICL achieves comparable or even better results than many-shot ICL in most cases, while substantially reducing the cost.
These results demonstrate the robustness of our method to decoding algorithms.

\section{Prompt Engineering for Cheat-Sheet Creation}
\label{app:expt_prompt}
In addition to the prompt presented in Appendix~\ref{app:cheat_prompt}, we tested several prompt variants for cheat-sheet creation.
Table~\ref{tab:prompt_var_desc} describes each variant and provides the full prompt text.
The \textsc{Textbook} prompt aims to produce a more detailed summary of the demonstrations.
The \textsc{Textual Summary} and \textsc{Concise Instruction} prompts probe alternative phrasings for cheat-sheet construction.
\textsc{More Format Examples} uses the same \textsc{Cheat Sheet} prompt as in our main results but increases formatting exemplars to isolate the effect of extra formatting guidance.

Table~\ref{tab:prompt_var_results} reports the downstream performance of each variant.
Overall, these alternatives are comparable to the \textsc{Cheat Sheet} prompt but, on average, perform slightly worse.
While this pattern underscores the robustness of the cheat-sheet creation procedure, it also suggests that broader textbook-style coverage does not improve--and may slightly degrade--performance; similarly, increasing the number of formatting examples yields little additional benefit.
These findings indicate that LLMs attend most effectively to essential information when presented with concise cheat sheets.
Additionally, the comparable yet lowest performance of the \textsc{Concise Instruction} variant suggests that explicitly reiterating the cheat-sheet construction objective within the prompt remains important.
Qualitatively, we observed that the \textsc{Concise Instruction} and \textsc{Textual Summary} variants tended to produce simple rule lists rather than the more visually interpretable tables shown in Figures~\ref{fig:cheat_sheet_disamb_2}, \ref{fig:cheat_sheet_movie_2}, and \ref{fig:cheat_sheet_causal_2} in Appendix~\ref{app:example_cheat}.

\section{Example: Manual Modification of the Cheat Sheet}
\label{app:manual_revision}

\looseness=-1
As described in Section~\ref{sec:error}, we found that cheat-sheet ICL in Disambiguation QA often incorrectly relied on common sense when the answer should be ``ambiguous''.
Since the cheat sheet is human-interpretable, we could easily identify and remove the \colorbox{diffred!40}{orange section \textcolor{white}{$-$}} that encouraged using world knowledge, and add an explicit instruction in the \colorbox{diffgreen!40}{green section \textcolor{white}{$+$}}  not to use it, as shown below.
This simple modification of the cheat sheet improved accuracy from 87.0 to 89.7.
The complete cheat sheet is provided in Appendix~\ref{app:example_cheat}.

\begin{tcolorbox}[title=Manual Modification of Cheat Sheet, colback=white, colframe=black!70, boxrule=0.5pt, top=0pt, bottom=0pt, left=0pt, right=0pt, arc=0.8mm]
...\\
\#\# 6. **Ambiguity Checklist**\\
- Both antecedents are grammatically possible.\\
- Both antecedents are logically possible.
\begin{tcolorbox}[
    colback=diffred!40,
    colframe=white,
    boxrule=0pt,
    leftrule=0pt, rightrule=0pt, toprule=0pt, bottomrule=0pt,
    sharp corners,
    boxsep=0pt, left=0pt, right=0pt, top=0pt, bottom=0pt,
    width=\textwidth,
    enlarge left by=0mm, enlarge right by=0mm,
    before skip=3pt, after skip=3pt,
]
- No context or world knowledge tips the scale.~\textcolor{white}{$\bm{-}$}
\end{tcolorbox}
- If all above are true, **choose "Ambiguous"**.
\begin{tcolorbox}[
    colback=diffgreen!40,
    colframe=white,
    boxrule=0pt,
    leftrule=0pt, rightrule=0pt, toprule=0pt, bottomrule=0pt,
    sharp corners,
    boxsep=0pt, left=0pt, right=0pt, top=0pt, bottom=0pt,
    width=\textwidth,
    enlarge left by=0mm, enlarge right by=0mm,
    before skip=3pt, after skip=6pt,
]
- Consider only the information explicitly provided and do not take into account any world knowledge or common sense beyond the given context. \textcolor{white}{$\bm{+}$}
\end{tcolorbox}
...
\end{tcolorbox}

\section{Details of Demonstration Retrieval}
\label{app:retrieval}
Following \citet{bertsch-etal-2025-context}, we adopted the retrieval methods employed by \citet{gupta-etal-2023-coverage}, specifically BM25, Cosine, and Set-BSR.
We employed the Okapi BM25 algorithm as implemented in \texttt{rank\_bm25} for BM25.\footnote{\url{https://github.com/dorianbrown/rank_bm25}}
For Cosine, we used \texttt{sentence-transformers} library \citep{reimers-gurevych-2019-sentence} to generate contextualized sentence embeddings and compute the cosine similarity between each test input and candidate demonstration inputs.\footnote{\url{https://github.com/UKPLab/sentence-transformers}}
In particular, we followed prior work by using the \texttt{all-mpnet-base-v2} model for embedding generation.
For the Set-BSR approach, we adopted the publicly available implementation provided by \citet{gupta-etal-2023-coverage}.\footnote{\url{https://github.com/Shivanshu-Gupta/icl-coverage}}
Following their paper, we used the \texttt{deberta-large-mnli} model.

\section{Effect of Nondeterminism in Proprietary Models}
\label{app:proprietary_var}
As noted in Appendix~\ref{app:decoding}, proprietary models are often nondeterministic and can produce different outputs even when the temperature is fixed at 0.
To quantify the impact of this nondeterminism on cheat-sheet construction and downstream performance, we ran each proprietary model three times under an identical decoding configuration with a fixed random seed; we report the resulting standard deviation in Table~\ref{tab:proprietary_var}.
While the magnitude of the effect is task-dependent, model nondeterminism has a modest effect on accuracy.
Importantly, although our method adds an additional cheat-sheet-creation stage that could, in principle, amplify variance, the observed standard deviation remains comparable to that of conventional few-shot and many-shot ICL settings.

For completeness, all other scores reported in this paper are averaged over three independent runs with different random seeds, comprising cheat-sheet creation followed by ICL.
Across runs, we vary the seed that controls demonstration shuffling, thereby accounting for both model-side nondeterminism and data-ordering effects.

\section{Example: Cheat Sheets}
\label{app:example_cheat}

We present examples of the cheat sheets that we have obtained in Figures~\ref{fig:cheat_sheet_disamb_1}--\ref{fig:cheat_sheet_causal_2}.

\input{figures/cheat_sheet_disamb_1}
\input{figures/cheat_sheet_disamb_2}
\input{figures/cheat_sheet_movie_1}
\input{figures/cheat_sheet_movie_2}
\input{figures/cheat_sheet_causal_1}
\input{figures/cheat_sheet_causal_2}

\end{document}

%% file: tables/retrieve.tex
\begin{table}[t]
\centering
\begin{adjustbox}{max width=1.0\columnwidth}
\begin{tabular}{lcc}
\toprule
 & \textbf{Accuracy} $\uparrow$ & \textbf{Input Token Length} $\downarrow$ \\
\midrule
8-shot & 87.1 & 2,334 \\
(150 or 100)-shot & \textbf{91.0} & 42,461 \\
\midrule
BM25 & 86.9 & \textbf{2,024} \\
Cosine & 89.1 & 2,294 \\
Set-BSR & 89.0 & 2,329 \\
\midrule
Cheat-sheet & \underline{90.0} & \underline{2,036} \\
\bottomrule
\end{tabular}
\end{adjustbox}
\caption{
Comparison with demonstration retrieval.
The scores are averaged across the eight BBH tasks.
\textbf{Bold} and \underline{underline} denote the best and second-best.
}
\label{tab:retrieve}
\end{table}

%% file: tables/acad_gemini.tex
\begin{table}[t]
\centering
\begin{adjustbox}{max width=1.0\columnwidth}
\begin{tabular}{lccc}
\toprule
 & \textbf{MATH500} & \textbf{GSM8K} & \textbf{GPQA} \\
\midrule
8-shot & 86.3 & 94.5 & 57.1 \\
(400 or 250)-shot & 82.3 & 95.1 & 56.7 \\
\midrule
Cheat-sheet & 88.0 & 94.5 & 60.4 \\
\bottomrule
\end{tabular}
\end{adjustbox}
\caption{
Performance of \texttt{gemini-1.5-pro-002} on academic tasks.
400-shot denotes using all available demonstrations on MATH500 and GSM8K, whereas 250-shot denotes the same on GPQA.
Scores are accuracies averaged over three runs.
The input token lengths match those shown in Figure~\ref{fig:acad}.
}
\label{tab:gemini_academic}
\end{table}

%% file: tables/cost.tex
\begin{table*}[t]
\centering
\begin{adjustbox}{max width=1.0\textwidth}
\begin{tabular}{lcccc}
\toprule
 & \multicolumn{2}{c}{\textbf{Avg. Token Length}} & \textbf{Cost for Input} & \textbf{Wall-Clock Time} \\
  & Input & Output & (USD) & (s) \\
\midrule
8-shot & 1,277 & 153 & 0.064 & 158.42 \\
150-shot & 23,921 & 144 & 1.196 & 287.52 \\
\midrule
Cheat-sheet & 1,306 & 155 & 0.065 & 158.71 \\
\bottomrule
\end{tabular}
\end{adjustbox}
\caption{
Monetary cost and wall-clock time for processing the test set of the Boolean Expressions task.
}
\label{tab:costs}
\end{table*}

%% file: tables/prompt_var_desc.tex
\begin{table*}[t]
\centering
\begin{adjustbox}{max width=0.98\textwidth}
\begin{tabular}{P{1.0\textwidth}}
    \toprule

    \PromptPair{\textbf{\textsc{Textbook}:} A prompt to produce a detailed, textbook-style overview. Similar in spirit to a cheat sheet, but aimed at covering broader task-relevant knowledge.}
    {Create a textbook based on the examples below. You will be asked to answer questions similar to these examples during the test, without being allowed to refer to the examples at that time. Your task here is to make a textbook that will help you answer such problems correctly. First, carefully read the examples below and identify the knowledge or reasoning steps required to answer similar questions correctly.\string\n\string\n\{$\hat{\D}_n$\}\string\n\string\nNow, create a textbook that thoroughly describes the knowledge or reasoning steps needed to answer similar questions correctly.} \\
    \midrule

    \PromptPair{\textbf{\textsc{Textual Summary}:} A variation of the cheat-sheet prompt that replaces only the term ``cheat sheet'' with ``textual summary''.}
    {Create a textual summary based on the examples below. You will be asked to answer questions similar to these examples during the test, without being allowed to refer to the examples at that time. Your task here is to make a textual summary that will help you answer such problems correctly. First, carefully read the examples below and identify which ones you find most difficult to answer.\string\n\string\n\{$\hat{\D}_n$\}\string\n\string\nNow, create a textual summary to help you solve the difficult examples. Exclude any content that is easy for you, and only include specific, detailed points to address the challenging ones.} \\
    \midrule

    \PromptPair{\textbf{\textsc{Concise Instruction}:} A more concise version of the cheat-sheet prompt.}
    {You will be asked to answer questions similar to the examples below, but you will not be allowed to refer to the examples during the test. First, carefully read the examples below and identify which ones you find most difficult to answer correctly.\string\n\string\n\{$\hat{\D}_n$\}\string\n\string\nNow, create a cheat sheet to help you address the difficult ones. Exclude any content that is easy for you, and include only specific, detailed points to address the difficult ones.} \\
    \midrule

    \PromptPair{\textbf{\textsc{More Format Examples}:} The prompt is unchanged; only the number of format examples appended to the output cheat sheet increases from two to eight.}
    {No change to the cheat-sheet prompt; simply increase $\hat{\D}_2$ in Eq.~\eqref{eq:cs-icl} to $\hat{\D}_8$.} \\

    \bottomrule
  \end{tabular}
\end{adjustbox}
\caption{Descriptions of tested prompt variants and complete prompt text.}
\label{tab:prompt_var_desc}
\end{table*}

%% file: tables/prompt_var_results.tex
\begin{table*}[t]
\centering
\begin{adjustbox}{max width=1.0\textwidth}
\begin{tabular}{lccccccccc}
\toprule
 & \textbf{Bool} & \textbf{Causal} & \textbf{DisambQA} & \textbf{Geo} & \textbf{Movie} & \textbf{Translation} & \textbf{Sports} & \textbf{Word} & \textbf{Avg.} \\
\midrule
\textsc{Textbook} & 100.0 & 66.3 & 86.0 & 95.3 & 90.0 & 77.0 & 99.0 & 97.3 & 88.9 \\
\textsc{Textual Summary} & 100.0 & 70.9 & 84.7 & 94.7 & 92.3 & 74.7 & 98.0 & 97.0 & 89.0 \\
\textsc{Concise Instruction} & 100.0 & 65.5 & 88.0 & 92.3 & 91.7 & 77.0 & 97.3 & 97.3 & 88.7 \\
\textsc{More Format Examples} & 100.0 & 67.4 & 88.7 & 93.3 & 93.3 & 75.7 & 99.7 & 98.7 & 89.6 \\
\textsc{Cheat Sheet} & 100.0 & 70.9 & 87.0 & 95.3 & 93.7 & 76.3 & 98.7 & 98.0 & 90.0 \\
\bottomrule
\end{tabular}
\end{adjustbox}
\caption{
Performance of GPT-4.1 on BBH for each prompt variant.
Task names are abbreviated; scores are accuracies averaged over three runs.
}
\label{tab:prompt_var_results}
\end{table*}

%% file: tables/proprietary_var.tex
\begin{table*}[t]
\centering
\begin{adjustbox}{max width=1.0\textwidth}
\begin{tabular}{lcccccccc}
\toprule
 & \textbf{Bool} & \textbf{Causal} & \textbf{DisambQA} & \textbf{Geo} & \textbf{Movie} & \textbf{Translation} & \textbf{Sports} & \textbf{Word} \\
\midrule
8-shot & 0.00 & 1.33 & 1.15 & 1.53 & 0.58 & 2.31 & 1.00 & 1.53 \\
(150 or 100)-shot & 0.00 & 0.66 & 0.58 & 0.58 & 0.00 & 1.53 & 0.58 & 2.65 \\
Cheat-sheet & 0.00 & 1.33 & 1.00 & 1.53 & 1.15 & 1.53 & 0.58 & 2.89 \\
\bottomrule
\end{tabular}
\end{adjustbox}
\caption{
Standard deviation of GPT-4.1 performance on BBH with a fixed random seed.
Task names are abbreviated; each value is the standard deviation computed over three runs with the same seed.
}
\label{tab:proprietary_var}
\end{table*}

%% file: figures/cheat_sheet_disamb_1.tex
\begin{figure*}[t]
\centering
\begin{tcolorbox}[title=Cheat Sheet for Disambiguation QA (1/2), colback=gray!10, colframe=black!70, boxrule=0.5pt, top=0pt, bottom=0pt, left=0pt, right=0pt, arc=0.8mm, width=1.0\textwidth]
\scriptsize
\begin{lstlisting}
---

# Pronoun Antecedent Cheat Sheet (for Difficult Cases)

## 1. **General Reasoning Steps**
- **Identify all possible antecedents** for the pronoun.
- **Substitute each antecedent** into the sentence to see if it makes sense.
- **Consider the context and logic** of the sentence: Who is likely to perform the action or possess the attribute?
- **Check for grammatical cues**: number (singular/plural), gender, and role in the sentence.
- **If both options are equally plausible and the sentence gives no extra clues, mark as ambiguous.**

---

## 2. **Common Patterns and How to Resolve Them**

### A. **"X told Y that [pronoun]..."**
- Usually, the pronoun refers to **Y** if the information is about Y (e.g., advice, diagnosis, payment).
- If the information is about X's own actions or status, it refers to **X**.
- **Tip:** Would it make sense for X to inform Y about Y's own actions? Usually not, unless it's advice or a warning.

### B. **"X did something to Y because [pronoun]..."**
- The pronoun can refer to either X or Y.
- **Test both:** Substitute both and see which makes more logical sense.
- If both are plausible, **mark as ambiguous**.

### C. **"X and Y discuss [pronoun]'s Z"**
- If both X and Y could logically possess Z, and the sentence gives no further context, **mark as ambiguous**.
- If only one is likely to possess Z (e.g., "culinary training" is more likely the chef's), pick that one.

### D. **"X called Y and asked [pronoun] to do Z"**
- The pronoun usually refers to **Y** (the person being asked to do something).
- If it would be odd for X to ask themselves, it's almost always Y.

### E. **"X met with Y at [pronoun]'s office"**
- If both X and Y could be the owner of the office, and the sentence gives no clue, **mark as ambiguous**.
- If only one is plausible (e.g., meeting a director at the director's office), pick that one.

### F. **"X did something with Y because [pronoun] [verb/attribute]"**
- If the verb/attribute fits both X and Y, and both are plausible, **mark as ambiguous**.
- If only one makes sense (e.g., "focuses on code" fits developer, not writer), pick that one.

### G. **Possessive Constructions ("the writer and [pronoun] friends")**
- The possessive pronoun almost always refers to the first noun ("the writer and her friends" = the writer's friends).
- If the pronoun could refer to more than one noun, but only one makes sense, pick that one.

---
\end{lstlisting}
\end{tcolorbox}
\caption{
An example of cheat sheet generated for Disambiguation QA.
This is the first half of the cheat sheet.
}
\label{fig:cheat_sheet_disamb_1}
\end{figure*}

%% file: figures/cheat_sheet_disamb_2.tex
\begin{figure*}[t]
\centering
\begin{tcolorbox}[title=Cheat Sheet for Disambiguation QA (2/2), colback=gray!10, colframe=black!70, boxrule=0.5pt, top=0pt, bottom=0pt, left=0pt, right=0pt, arc=0.8mm, width=1.0\textwidth]
\scriptsize
\begin{lstlisting}
## 3. **Ambiguity Triggers**
- If both antecedents are equally plausible and the sentence gives no further context, **choose "Ambiguous"**.
- Watch for sentences where both X and Y could have performed the action, received the attribute, or owned the object.

---

## 4. **Special Cues**
- **Gender/Number Agreement:** Make sure the pronoun matches the possible antecedent in gender and number.
- **Role/Profession:** Sometimes, the profession or role makes one antecedent more likely (e.g., only a scientist needs a lab assistant).
- **Typical Scenarios:** Use real-world logic (e.g., a mechanic calls a customer about the customer's car, not their own).

---

## 5. **Quick Reference Table**

| Structure                                      | Most Likely Antecedent | When Ambiguous?                |
|------------------------------------------------|-----------------------|-------------------------------|
| X told Y that [pronoun]...                     | Y (if advice/info)    | If both could be true         |
| X did Y because [pronoun]...                   | X or Y (test both)    | If both make sense            |
| X and Y discuss [pronoun]'s Z                  | Context-dependent     | If both could own Z           |
| X called Y and asked [pronoun] to do Z         | Y                     | If both could be asked        |
| X met with Y at [pronoun]'s office             | Context-dependent     | If both could own office      |
| X did Y because [pronoun] [verb/attribute]     | Context-dependent     | If both fit                   |
| The writer and [pronoun] friends               | The writer            | If only one makes sense       |

---

## 6. **Ambiguity Checklist**
- Both antecedents are grammatically possible.
- Both antecedents are logically possible.
- No context or world knowledge tips the scale.
- If all above are true, **choose "Ambiguous"**.

---

**Use this sheet to reason through each step, especially when both antecedents seem possible!**
\end{lstlisting}
\end{tcolorbox}
\caption{
An example of cheat sheet generated for Disambiguation QA.
This is the second half of the cheat sheet.
}
\label{fig:cheat_sheet_disamb_2}
\end{figure*}

%% file: figures/cheat_sheet_movie_1.tex
\begin{figure*}[t]
\centering
\begin{tcolorbox}[title=Cheat Sheet for Movie Recommendation (1/2), colback=gray!10, colframe=black!70, boxrule=0.5pt, top=0pt, bottom=0pt, left=0pt, right=0pt, arc=0.8mm, width=1.0\textwidth]
\scriptsize
\begin{lstlisting}
---

## CHEAT SHEET: "Find a Movie Similar To..." (Difficult Cases)

### 1. **Identify the Main Pattern**
- **Era/Decade:** Most correct answers are from the same decade as the given movies (often 1990s, sometimes 1980s or 2000s).
- **Genre:** Match the dominant genres (e.g., action, adventure, drama, crime, sci-fi, animation, family).
- **Fame/Recognition:** The answer is almost always a well-known, mainstream, or critically acclaimed film.
- **Tone/Style:** If the given movies are light-hearted, family-friendly, or epic, the answer should match that tone.

---

### 2. **Common Movie Pools**
- **1990s Hollywood Blockbusters:** The Shawshank Redemption, Forrest Gump, Pulp Fiction, Braveheart, Schindler's List, The Fugitive, Dances with Wolves, The Lion King, Toy Story, Pretty Woman, Apollo 13, Independence Day, Jurassic Park, The Silence of the Lambs, Batman, The Mask, Get Shorty, The Usual Suspects, Crimson Tide, Fargo, Goodfellas, LA Confidential, Philadelphia, True Lies, Heat, Seven, Forrest Gump, The Matrix, Gladiator, Gattaca, Inception.
- **Classic Animation/Family:** The Lion King, Aladdin, Toy Story, Beauty and the Beast, Pinocchio, The Jungle Book, The Wizard of Oz, Snow White, Fantasia.
- **Classic Sci-Fi/Adventure:** Star Wars (original trilogy), Raiders of the Lost Ark, The Terminator, Back to the Future, The Matrix, Terminator 2, Independence Day, Stargate, The Fifth Element.
- **Crime/Drama/Thriller:** Pulp Fiction, The Shawshank Redemption, The Usual Suspects, Goodfellas, LA Confidential, Fargo, Seven, The Silence of the Lambs, Heat, Get Shorty, Crimson Tide, The Fugitive.

---

### 3. **How to Eliminate Wrong Options**
- **Obscure/Unfamiliar Titles:** If you don't recognize a title, it's probably not the answer.
- **Genre Mismatch:** If the option is a comedy and the given movies are all dramas, eliminate it.
- **Era Mismatch:** If the option is from a much earlier or later decade, eliminate it.
- **Foreign/Indie/Low-Profile:** If the option is a foreign film or a low-profile indie, and the given movies are Hollywood blockbusters, eliminate it.

---

### 4. **Special Patterns & Tricky Cases**
- **Franchise/Sequel/Director Overlap:** If the given movies are from a franchise or share a director, and an option is from the same franchise/director, it's likely the answer.
- **Animation Among Live-Action:** If the list includes both animation and live-action, the answer can be either, but it must be a *famous* one from the same era.
- **Mix of Genres:** If the given movies are a mix (e.g., action, drama, animation), the answer is usually a famous, mainstream film from the same period, even if the genre is not an exact match.
- **Critical Acclaim:** If all the given movies are Oscar winners/nominees or have high critical acclaim, the answer should be similarly acclaimed.

---

### 5. **When Multiple Options Seem Plausible**
- **Choose the Most Famous:** Go with the most universally recognized title.
- **Check for Cast/Director Overlap:** Sometimes, the answer shares actors or directors with the given movies.
- **Check for Cultural Impact:** The answer should have a similar level of cultural impact as the given movies.

---
\end{lstlisting}
\end{tcolorbox}
\caption{
An example of cheat sheet generated for Movie Recommendation.
This is the first half of the cheat sheet.
}
\label{fig:cheat_sheet_movie_1}
\end{figure*}

%% file: figures/cheat_sheet_movie_2.tex
\begin{figure*}[t]
\centering
\begin{tcolorbox}[title=Cheat Sheet for Movie Recommendation (2/2), colback=gray!10, colframe=black!70, boxrule=0.5pt, top=0pt, bottom=0pt, left=0pt, right=0pt, arc=0.8mm, width=1.0\textwidth]
\scriptsize
\begin{lstlisting}
### 6. **Examples of Subtle Connections**
- **Animation/Family:** If the list includes The Lion King, Toy Story, Aladdin, the answer is likely another 90s animation (e.g., Beauty and the Beast, Pinocchio).
- **Crime/Drama:** If the list includes Pulp Fiction, The Usual Suspects, The Shawshank Redemption, the answer is likely another 90s crime/drama (e.g., Get Shorty, LA Confidential, Seven).
- **Action/Adventure/Sci-Fi:** If the list includes Star Wars, The Matrix, Raiders of the Lost Ark, the answer is likely another big-budget action/sci-fi/adventure from the same era (e.g., Terminator 2, Independence Day, The Fifth Element).
- **Historical/Epic Drama:** If the list includes Braveheart, Schindler's List, Dances with Wolves, the answer is likely another 90s historical/epic drama (e.g., Apollo 13, Gladiator, Philadelphia).

---

### 7. **If Stuck: Default to These Titles**
If you're unsure, and the options include any of these, they are *very* often correct:
- The Shawshank Redemption
- Forrest Gump
- Braveheart
- Dances with Wolves
- The Fugitive
- Pulp Fiction
- The Lion King
- Independence Day
- Terminator 2: Judgment Day
- Get Shorty
- LA Confidential
- Gladiator
- Raiders of the Lost Ark
- Toy Story
- The Matrix

---

### 8. **Quick Reference Table**

| Given List Features                | Look for Option Like...         |
|------------------------------------|---------------------------------|
| 90s Hollywood, drama/crime         | Shawshank, Pulp Fiction, Usual Suspects, Get Shorty, LA Confidential, Seven, Forrest Gump |
| 90s Hollywood, action/adventure    | The Fugitive, Terminator 2, Independence Day, Gladiator, True Lies, Heat, Braveheart |
| 90s Animation/Family               | The Lion King, Aladdin, Toy Story, Beauty and the Beast, Pinocchio |
| 80s/90s Sci-Fi/Adventure           | Star Wars, Raiders, The Matrix, Terminator 2, Back to the Future, Fifth Element |
| Historical/Epic                    | Braveheart, Dances with Wolves, Apollo 13, Gladiator, Schindler's List |
| Mix of genres, all famous          | Pick the most famous, acclaimed, or era-matching option |

---

### 9. **Red Flags for Wrong Answers**
- Obscure, foreign, or recent indie films
- Comedies when the list is all drama/thriller
- Animated films when the list is all live-action (unless the animation is a 90s classic)
- Movies from a much earlier or later decade

---

**REMEMBER:**  
When in doubt, match **era + genre + fame/impact**. If you see a 90s classic among the options and the list is 90s classics, pick it!

---

**End of Cheat Sheet**
\end{lstlisting}
\end{tcolorbox}
\caption{
An example of cheat sheet generated for Movie Recommendation.
This is the second half of the cheat sheet.
}
\label{fig:cheat_sheet_movie_2}
\end{figure*}

%% file: figures/cheat_sheet_causal_1.tex
\begin{figure*}[t]
\centering
\begin{tcolorbox}[title=Cheat Sheet for Causal Judgement (1/2), colback=gray!10, colframe=black!70, boxrule=0.5pt, top=0pt, bottom=0pt, left=0pt, right=0pt, arc=0.8mm, width=1.0\textwidth]
\scriptsize
\begin{lstlisting}
- Multiple necessary conditions (e.g., two gardeners/fertilizers, two wires, two people logging in, etc.)
- Overdetermination (multiple sufficient causes)
- Policy/Norm violations vs. permitted actions
- Side effects and intention
- Chains of causation (proximate vs. remote causes)
- "Because" questions with multiple sufficient conditions

---

## CHEAT SHEET: CAUSATION & INTENTION

### 1. **Multiple Necessary Conditions (Joint Causation)**
- **If an outcome only happens when two (or more) actions/conditions occur together,** each action is *necessary* but not *sufficient* alone.
    - **Typical person:** Usually says *No* to "Did X cause Y?" if X alone is not sufficient, unless X is the abnormal or rule-breaking action.
    - **Exception:** If X is the abnormal/forbidden action (e.g., red wire not supposed to touch battery), people may attribute causation to X.

#### **Example: Two Wires**
- Machine shorts only if both black and red wires touch battery.
    - Black wire is supposed to touch; red is not.
    - **Did black wire cause short?** -> **No** (normal/expected action)
    - **Did red wire cause short?** -> **Yes** (abnormal/unexpected action)

#### **Example: Two Gardeners/Fertilizers**
- Plants dry out only where both fertilizers are applied.
    - **Did Alex (A X200R) cause drying?** -> **No** (if only A X200R is used, no drying)
    - **Did Benni (B Y33R) cause drying?** -> **No** (if only B Y33R is used, no drying)
    - **Did Alex cause drying in beds with both?** -> **Yes** (if question is about the *combination* and Alex's action was necessary for the harmful combo)
    - **If Benni's action is abnormal (e.g., used wrong fertilizer), more likely to attribute causation to Benni.**

#### **Example: Two People Logging In**
- Deletion/email only happens if both are logged in.
    - **If one is violating policy and the other is not:**
        - **Violator:** *Yes*, caused the outcome.
        - **Permitted user:** *No*, did not cause the outcome.

---

### 2. **Overdetermination (Multiple Sufficient Causes)**
- **If either of two actions is sufficient to cause the outcome, and both occur:**
    - **Typical person:** Each action is seen as a cause.
    - **"Did X cause Y?"** -> **Yes** (if X alone would have been enough)
    - **"Did X cause Y because of Z?"** -> **No** (if Y would have happened anyway due to another sufficient cause)

#### **Example: Bridge Collapse**
- If either train alone is enough to collapse the bridge:
    - **Did Billy cause collapse?** -> **Yes**
- If both trains are needed:
    - **Did Billy cause collapse?** -> **No**

#### **Example: Coffee Shop**
- If any one customer is enough for profit, and several order:
    - **Did Drew cause profit?**
        - If others would have ordered anyway: **No**
        - If Drew was the only one: **Yes**

---

### 3. **Policy/Norm Violations vs. Permitted Actions**
- **If two people act, but only one violates a rule:**
    - **Violator:** *Yes*, caused the outcome.
    - **Permitted actor:** *No*, did not cause the outcome.

#### **Example: Computer Crash**
- Jane (permitted) logs in, Lauren (violator) logs in, crash occurs.
    - **Did Jane cause crash?** -> **No**
    - **Did Lauren cause crash?** -> **Yes**

---
\end{lstlisting}
\end{tcolorbox}
\caption{
An example of cheat sheet generated for Causal Judgement.
This is the first half of the cheat sheet.
}
\label{fig:cheat_sheet_causal_1}
\end{figure*}

%% file: figures/cheat_sheet_causal_2.tex
\begin{figure*}[t]
\centering
\begin{tcolorbox}[title=Cheat Sheet for Causal Judgement (2/2), colback=gray!10, colframe=black!70, boxrule=0.5pt, top=0pt, bottom=0pt, left=0pt, right=0pt, arc=0.8mm, width=1.0\textwidth]
\scriptsize
\begin{lstlisting}
### 4. **Side Effects and Intention**
- **If someone foresees but does not care about a side effect:**
    - **Harmful side effect:** *Yes*, intentionally caused (Knobe effect).
    - **Helpful side effect:** *No*, not intentionally caused.

#### **Example: CEO/environment**
- CEO knows program will harm environment, doesn't care, proceeds.
    - **Did CEO intentionally harm environment?** -> **Yes**
- CEO knows program will help environment, doesn't care, proceeds.
    - **Did CEO intentionally help environment?** -> **No**

#### **Example: Hunter/Eagle**
- Hunter knows gunshot will scare eagle, doesn't care, shoots deer.
    - **Did hunter intentionally scare eagle?** -> **No**

---

### 5. **Chains of Causation (Proximate vs. Remote)**
- **If an immediate cause interrupts a chain (e.g., nurse's error causes death before cancer):**
    - **Immediate cause (nurse's error):** *Yes*, caused death.
    - **Underlying cause (cancer, asbestos):** *Yes*, if question is about "premature death" or "set in motion" the chain.
    - **Job/relocation:** *No*, if immediate cause is unrelated (e.g., medication error).

---

### 6. **"Because" Questions with Multiple Sufficient Conditions**
- **If outcome would have happened anyway due to another sufficient condition:**
    - **"Did Y happen because of X?"** -> **No**
    - **"Did Y happen because of X and Z?"** -> **Yes** (if both are necessary)
    - **If X is not necessary, answer is No.**

#### **Example: Free Sample**
- Laurie gets sample if she bought beans or is on email list.
    - She qualifies both ways.
    - **Did she get sample because she changed subscription?** -> **No** (already qualified)
    - **Did she get sample because she did not unsubscribe?** -> **Yes** (if her continued subscription was necessary for eligibility)

---

### 7. **Grading on a Curve / Competitive Scenarios**
- **If a person's action directly blocks another from achieving a result (e.g., last A in a curve):**
    - **Did X cause Y's failure?** -> **Yes** (if X's action was necessary for Y's failure)

---

### 8. **Intentionality and Accidents**
- **If outcome is due to accident/lack of control (e.g., hand slips, dart wobbles):**
    - **Did X intentionally do Y?** -> **No** (even if X wanted Y, lack of control means not intentional)

---

## **Quick Reference Table**

| Scenario Type                        | Typical Person's Answer |
|--------------------------------------|-------------------------|
| Both actions needed (joint cause)    | No (unless abnormal)    |
| Either action sufficient (overdet.)  | Yes                     |
| Policy violator vs. permitted        | Violator: Yes; Permitted: No |
| Side effect (harmful, foreseen)      | Yes (intentional)       |
| Side effect (helpful, foreseen)      | No (not intentional)    |
| Immediate vs. remote cause           | Immediate: Yes; Remote: Yes if chain is relevant, No if not |
| "Because" with multiple sufficients  | No                      |
| Grading on a curve                   | Yes                     |
| Accidental outcome                   | No (not intentional)    |

---

**TIP:**  
- Always ask: Was the action necessary and/or sufficient for the outcome?  
- Was the action abnormal or a violation?  
- Was the outcome intended, foreseen, or a side effect?  
- Would the outcome have happened anyway without this action?

---

**Use this sheet to reason through the tricky causation and intention questions!**
\end{lstlisting}
\end{tcolorbox}
\caption{
An example of cheat sheet generated for Causal Judgement.
This is the second half of the cheat sheet.
}
\label{fig:cheat_sheet_causal_2}
\end{figure*}